\documentclass[10pt, a4paper]{article}

\usepackage[final]{lrec-coling2024} 

\usepackage{times}
\usepackage{latexsym}
\usepackage{graphicx} 
\usepackage{amsmath}
\usepackage{amssymb}
\usepackage{amsfonts}
\usepackage{ascmac}
\usepackage{url}

\title{User-Specific Dialogue Generation with User Profile-Aware Pre-Training Model and Parameter-Efficient Fine-Tuning}

\name{Atsushi Otsuka, Kazuya Matsuo, Ryo Ishii, Narichika Nomoto, Hiroaki Sugiyama} 

\address{NTT Human Informatics Laboratories,NTT Corporation\\
         \{atsushi.otsuka, kazuya.matsuo, ryoct.ishii, narichika.nomoto, hiroaki.sugiyama\}@ntt.com\\}

\abstract{
This paper addresses user-specific dialogs. In contrast to previous research on personalized dialogue focused on achieving virtual user dialogue as defined by persona descriptions, user-specific dialogue aims to reproduce real-user dialogue beyond persona-based dialogue. Fine-tuning using the target user's dialogue history is an efficient learning method for a user-specific model. However, it is prone to overfitting and model destruction due to the small amount of data. Therefore, we propose a learning method for user-specific models by combining parameter-efficient fine-tuning with a pre-trained dialogue model that includes user profiles. Parameter-efficient fine-tuning adds a small number of parameters to the entire model, so even small amounts of training data can be trained efficiently and are robust to model destruction. In addition, the pre-trained model, which is learned by adding simple prompts for automatically inferred user profiles, can generate speech with enhanced knowledge of the user's profile, even when there is little training data during fine-tuning. In experiments, we compared the proposed model with large-language-model utterance generation using prompts containing users' personal information. Experiments reproducing real users' utterances revealed that the proposed model can generate utterances with higher reproducibility than the compared methods, even with a small model.
 \\ \newline \Keywords{dialogue model, real users, user profile, parameter-efficient fine-tuning} }

\begin{document}

\maketitleabstract

\section{Introduction}

Deep learning-based dialogue models have made significant progress over the years. They can now conduct conversations comparable to human-to-human dialogue on any given topic \cite{shuster2022blenderbot, roller-etal-2021-recipes, 10022973, shahriar2023lets}. With the improvement of chatbot performance, conversational AI has now been introduced in various services spanning the metaverse \cite{10099167} and gaming \cite{9782047} industries. Recent research has reported that it is possible to simulate the conversation and behavior of multiple agents using a large language model \cite{park2023generative}, where dialogue agents are given personalities and dialogue in accordance with their personalities. If these personalities could be extended to real people, it would be possible to instantly form a consensus among a large number of people, which is a task that has been impossible until now due to time and space constraints \cite{Toshima_ntt}. Therefore, in this paper, we aim to reproduce real users through a user-specific dialogue model.

Much of the research on giving personality to conversational AI has been conducted using the PERSONA-CHAT dataset \cite{zhang-etal-2018-personalizing}, which consists of several persona descriptions and dialogue histories. It evaluates the ability to generate consistent dialogues without contradictions while referring to persona descriptions. However, the dialogue histories in PERSONA-CHAT are generated through role-playing by crowd workers on the basis of persona descriptions and are not actual conversations between real people. Furthermore, the personas in the dataset are defined by only a few sentences, which limits the ability to reproduce knowledge or thoughts beyond the persona description. Thus, different approaches are required to realize a user-specific dialogue model that can reproduce real users.

Fine-tuning is a practical approach for generating utterances that replicate a specific character \cite{higashinaka-etal-2018-role, 10.1007/978-981-19-5538-9_20}. Dialogue histories must be collected in advance to reproduce a specific user's dialogue using fine-tuning. However, collecting a large amount of dialogue data can be difficult from a privacy perspective, as such data may contain personal information. In previous research, the personality of a famous character was replicated by collecting the dialogue data of users who role-played that character, but this approach is limited to cases where the personality being collected is well-known and has role-playable characteristics. Therefore, we need an approach that can learn real user personalities even with small amounts of data that can be collected during short dialogues.

In recent years, many studies have reported using large language models (LLM) to give dialogue models individuality \cite{kasahara-etal-2022-building, ramirez2023controlling}. In these studies, the personal characteristics of the interlocutor are described before the dialogue history is given to the language model so that utterances reflecting the described personality are generated from the language model. The imposition of individuality through prompts significantly affects personality-based utterance generation. However, the models are easily influenced by the content of the prompts and are not good at producing utterances about content not written in the prompts. In addition, LLMs have huge model parameters and require many hardware resources to have individual personalized models, even for fine-tuning with adapters such as LoRA \cite{DBLP:conf/iclr/HuSWALWWC22}.

In this paper, we aim to realize a dialogue model that reproduces the dialogue of many real users. To this end, we propose a model that combines a small pre-learning model with adapter-based fine-tuning. The pre-training model also includes simple prompts based on automatically inferred user profiles to generate utterances based on the user profile. Then, by fine-tuning with LoRA, even a small model and a small amount of data can be used to generate utterances that reflect the individuality and characteristics of real users.

\noindent
Our main contributions are as follows.
\begin{itemize}
\item We develop a user-specific dialogue model reproducing real users' personalities by combining a small pre-training model that considers user profiles with LoRA fine-tuning.
\item We train over 9,000 user-specific models from 300 users with multiple pre-training and fine-tuning combinations and use these models to conduct a comparative experiment.
\item We compare the proposed model and LLMs, which includes user profile information in the prompts. The experiment results show that our model, even on a small scale, can generate utterances closer to those of real users than utterance generation done using LLMs with prompts.
\end{itemize}

\section{Related Work}

\paragraph{Personalized dialogue dataset} The PERSONA-CHAT dataset is one of the most widely used datasets for research on personalized dialogue, and it is also used as a dataset for competitions \cite{10.1007/978-3-030-29135-8_7}. In addition, it has been used as a basis for new datasets that include additional empathetic conversation data \cite{zhong-etal-2020-towards} and datasets created from website data \cite{mazare-etal-2018-training}. Previous studies have also examined manipulating PERSONA-CHAT data to improve data quality \cite{cao-etal-2022-model}.

Datasets in past studies have been constructed by collecting persona descriptions that can be represented as key-value pairs such as age and gender \cite{ijcai2018p595}. Zheng et al. \cite{zheng2020personalized, zheng2020aai} collected large-scale dialogue data with user profiles by estimating user information from a Chinese microblogging service and proposed a dialogue model pre-trained with the collected data. The methods in these studies are very similar to our own. The estimated user profiles are concatenated into the dialogue context as an input sequence in our method, which means it can be used without modifying a general language model.

\paragraph{Personalized dialogue generation}

Deep learning-based utterance generation models have been successful for open-domain conversations. 
Previous studies have added user embeddings to the utterance generation model to generate personalized dialogues \cite{li-etal-2016-persona, ijcai2017p521}, and several methods generate consistent utterances based on reading persona descriptions \cite{song-etal-2021-bob, 10.1007/978-3-030-95408-6_15, 9956081}. 
Other methods reference persona information in line with the content of the dialogue \cite{huang2023personalized} or refine the dialogue context to be input to the dialogue model, leaving only the context relevant to the persona \cite{zhong-etal-2022-less}. 
Prior research has reported methods that infer latent persona information from dialogue history \cite{cho-etal-2022-personalized} or that store long-term memories as persona descriptions \cite{xu-etal-2022-long}. A metric to assess whether utterances match the persona characteristics has also been proposed \cite{miyazaki-etal-2021-fundamental}.

We aim to achieve dialogues that reproduce real users, not description-defined personas, 
and adopt a training approach for user-specific models through fine-tuning rather than referring to descriptions.

\begin{figure*}[t]
\centering
\includegraphics[width =.8\textwidth]{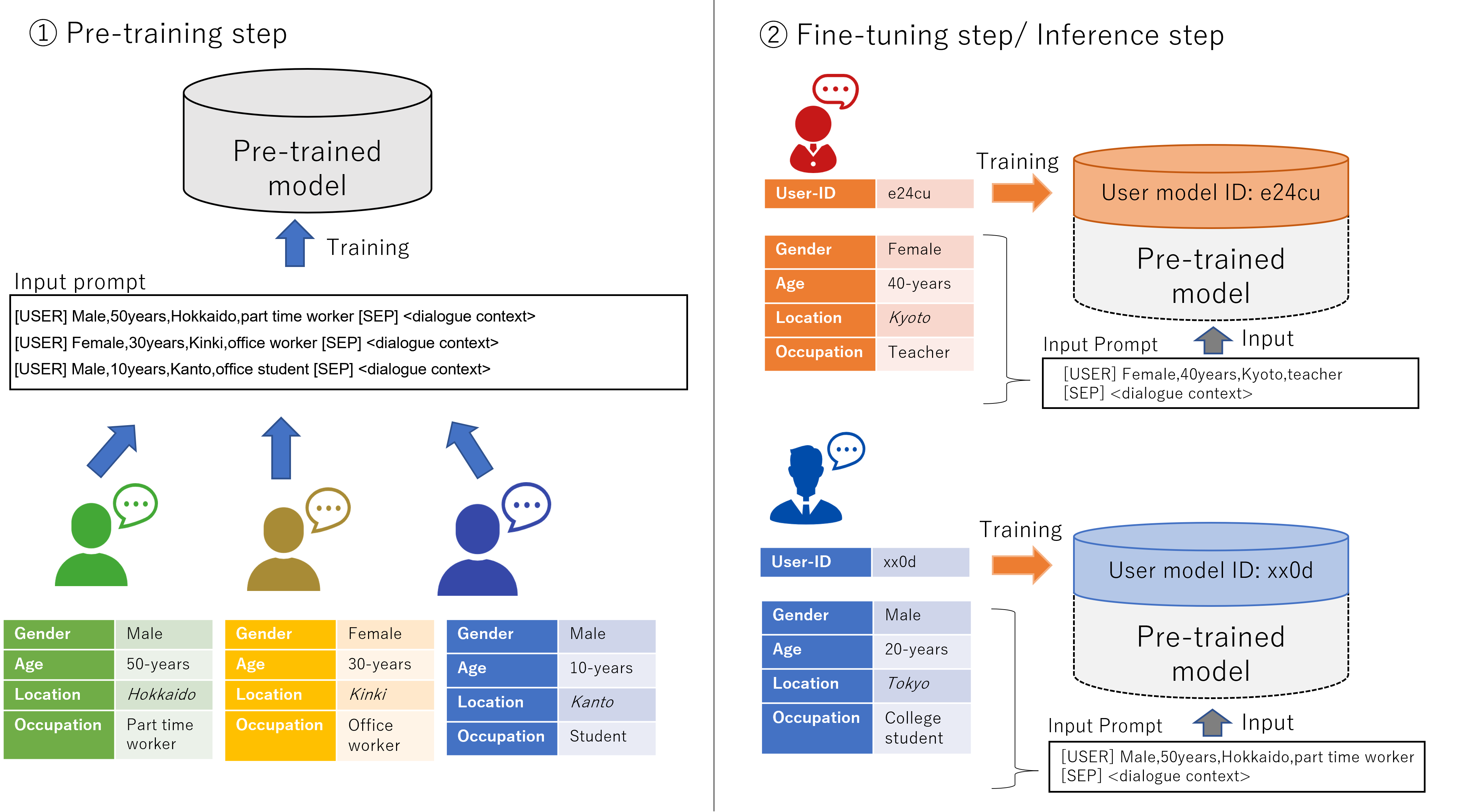}
\vspace{-0.5em}
\caption{\footnotesize Overview of proposed method. First, we train pre-training model using user profiles and dialogue contexts from many users. Next, we train user-specific model for users with user ID.}
\label{fig:over}
\end{figure*}

\section{Method}

\subsection{Task definition} \label{subsec:task}
Our task for generating responses given input consisting of dialogue context, persona, and user ID information can be formulated as

\begin{equation}
Y = \underset{Y} {\operatorname{argmax}} P(Y|X,U,I),
\end{equation}

where $X=\{x_0,...,x_n\}$ is a dialogue context, $U=\{u_{p0},...,u_{pm}\}$ is a user profile, $Y=\{y_0,...,y_l\}$ is a generated response, and $I$ denotes a user ID. $x$ and $y$ are tokens for input to the deep learning model. $u$ represents the user profile items (e.g., gender and age). The user ID is a unique and random string.
Note that we are working on reproducing a specific person's dialogue, which is why we require a user ID to identify the person.

\subsection{Overview}
An overview of our personalized dialogue generation model is shown in Fig. \ref{fig:over}. The first step is to train a pre-trained model using social dialogue data such as social networking site (SNS) reply pairs. At this time, we infer the profile of the SNS user and include the inferred profile in the input prompts. The next step is to train a model of a real user. During the fine-tuning step, only the user-specific model corresponding to the user ID is trained, and the parameters of the pre-trained model are not updated. The input prompts are entered in the same format as for the pre-training.


\subsection{Inferring user profile for pre-training} \label{subsec:markov}

Large language models (LLMs) describe information that is characteristic of the user in prompts so that utterances are based on the content of the prompts. However, for smaller models, prompting has a limited effect. Therefore, we pre-trained the dialogue model with prompts that included the dialogue context and user profile information. In a pre-trained model that does not include user profiles, the model is learned by dialogue from many users who have various backgrounds on SNSs. In comparison, the proposed model learns the user's profile information as a condition for utterance generation in addition to the dialogue history. Even with the same dialogue history, different utterances can be generated by changing the user's profile.

Since training a pre-trained model requires a huge amount of dialogue data, we use SNS replies as dialogue data for model training. However, obtaining the uniform user profiles needed for model training from SNSs is difficult. Therefore, we use a method for estimating user profiles from text posted on SNSs. We obtain a large number of user attributes using a Markov logic-based method \cite{richardson06markov} for estimating user profiles from SNSs developed by Hirano et al. \cite{Hirano2013}. They proposed a technique that estimates basic profiles such as age and place of residence from text posted on microblogs like X (Twitter) and reported that 150 pieces of text are needed to estimate user attributes. Therefore, we used a method of dividing the user's posted text for one year into 150 units, estimating the user's profile in each, and taking a majority vote.

Our user profiles are simplified compared with LLM prompts, so we cannot flexibly describe every user image into prompts. However, during model fine-tuning, prompts can include attributes tailored to the actual user's profile so that the prompts can supplement knowledge about that profile lacking in a small number of training data.
 
\subsection{Learning pre-trained model}\label{subsec:pre-train}

We train a generative language model based on the Transformer encoder-decoder model \cite{10.5555/3295222.3295349} during the pre-training step using dialogue data that has inferred user profiles. Our method inputs an estimated user profile as a string directly into the input prompt:
\begin{itembox}[l]{PSP: Persona Speaker Prompt}
\small
[USER]$u_{p1}$,$u_{p2}$,..,$u_{pm}$[SEP]$x_0$,$x_1$,..,$x_n$
\end{itembox}
where [USER] and [SEP] are special tokens. Our method can be used without depending on the model by including the user profile in the input prompts. Additionally, the high expressive power of the language model allows for robust handling of the description of user profiles, such as detailed age or geographical information.

The input prompt above includes the user profile of the speaker, and we call it a Persona Speaker Prompt (PSP). In reality, the content of speech may be subject to changes not only by the speaker but also by the conversational partner. We, therefore, propose an additional Persona Pair Prompt (PPP) that includes the speaker's profile and the conversational partner's profile in the input prompt.

\begin{itembox}[l]{PPP: Persona Pair Prompt}
\small
[USER1]$u_{p1}$,$u_{p2}$,..,$u_{pm}$[USER2]$u'_{p1}$,$u'_{p2}$,..,$u'_{pm}$

[SEP]$x_0$,$x_1$,..,$x_n$
\end{itembox}

The special tokens [USER1] represent the speaker's profile, and [USER2] represents the partner's profile.

\subsection{Parameter-efficient fine-tuning} \label{subsec:LoRA}

During the fine-tuning step, a personalized model for the user is trained using both the pre-trained model and the user's dialogue data. In this paper, we train user-specific models using parameter-efficient fine-tuning, which involves learning only a subset of the parameters of a pre-trained model rather than all. Reducing the number of parameters to be learned reduces the cost of operating many user models.

We use LoRA \cite{DBLP:conf/iclr/HuSWALWWC22} to perform fine-tuning, which focuses on the difference in updates to parameters. The output $h' \in \mathbb{R}^{d}$ of the forward layer of the fine-tuned model is calculated on the basis of the weight of pre-trained model $W \in \mathbb{R}^{d \times d}$ and input $x \in \mathbb{R}^{d}$ as:
\begin{equation}
h' = (W + \Delta W)x,
\end{equation} 
where $d$ denotes the dimension of the network and $\Delta W$ represents the difference in updated weights through fine-tuning, which can be further expressed as
\begin{equation}
\Delta W = BA,
\end{equation} 
where $B \in \mathbb{R}^{d \times r}$ and $A \in \mathbb{R}^{r \times d}$. Here, $r$ denotes the rank and is set to a small value for $d$. The number of learning parameters is $2dr$, much smaller than full fine-tuning. Since the pre-trained model weights $W$ are not updated, switching to a user-specific model becomes possible by modifying only $BA$ in accordance with the user ID.

\section{Experiments}

\subsection{Model training for experiments}

We first describe the comparative models. 
The experiment compares fine-tuning models and does not evaluate the pre-trained models themselves. 
This is because the pre-trained models do not allow for the input of the user ID $I$ required for the task we defined in \ref{subsec:task}.

\subsubsection{Dataset for pre-trained models} \label{subsec:dataset}

\begin{table}[t]
  \centering
  \scriptsize
  \begin{tabular}{lr|lr}
    \hline
    \multicolumn{2}{c|}{Age} & \multicolumn{2}{c}{Gender} \\ \hline
    20-years &  690,842,392 &Female & 557,836,904 \\
    30-years &  128,440,221 & Male & 441,809,462 \\ \cline{3-4}
    40-years &  103,304,429 &\multicolumn{2}{c}{Marriage} \\ \cline{3-4}
    10-years &  63,883,673  &Married& 147,192,421 \\
    60-years &  5,469,681 &&\\
    50-years &  4,171,332 &&\\ \hline
    \multicolumn{2}{c|}{Occupation} & \multicolumn{2}{c}{Location} \\ \hline
    Office worker &  405,076,925 &   \it{Kanto} & 538,096,358\\
    College student & 189,376,550 & \it{Kinki} & 272,929,741 \\
    Part-time worker &  158,019,739 & \it{Tokai} & 97,228,317 \\
    Unemployed &  58,781,115 & \it{Kyushu/Okinawa} &  13,084,462 \\
    Homemaker &  56,013,691 &  \it{Tohoku/Hokkaido} & 10,943,549 \\
    Business owner & 37,425,892 & \it{Chugoku/Shikoku} & 4,911,275 \\
    High school student &  7,480,769 & \it{Hokuriku} & 658,356  \\
    Association member &  5,347,385 & & \\
    Civil servant &  86,968 & & \\ \hline
  \end{tabular}
  \vspace{-1em}
  \caption {\footnotesize Inferred X (Twitter) user profiles. Note that unknown and other labels are excluded.}
  \label{table:users}
\end{table}

We used Twitter reply pairs to train the pre-trained models for our experiments and used the data creation method developed by Sugiyama et al. \cite{10022973}. A pre-training dataset was created with 1.3 billion reply pairs for the 2020-- 2021 period for 1 million users. Table \ref{table:users} shows the inferred user profile information. The training of the pre-training models with PPP and PSP contained input prompts that concatenated these user profiles. 

\subsubsection{Pre-trained models} 
\label{subsub:pretrain}
The following four pre-trained models were used in the experiments. All models features a Transformer encoder-decoder architecture with 222 M parameters.

\paragraph{Plain T5} This is a general Japanese language model \cite{10.5555/3455716.3455856}\footnote{\url{https://huggingface.co/sonoisa/t5-base-japanese}} known to be highly effective for fine-tuning with small amounts of data. 

\paragraph{Plain Dialogue} We trained a dialogue model using only the dialogue context from X (Twitter) reply pairs without user profile prompts.

\paragraph{PSP Dialogue} This pre-trained model trained by PSP includes the speaker's user profile and dialogue context described in \ref{subsec:pre-train} in the input prompts.

\paragraph{PPP Dialogue} This pre-trained model trained by PPP includes the user profiles and dialogue context of the speaker and the partner in the input prompts described in \ref{subsec:pre-train}.

\subsubsection{Fine-tuning training} \label{subsub:finetun}

For fine-tuning, we used a pre-trained model and a small amount of user dialogue data to train a user-specific dialogue model. We then compared the following fine-tuning methods in our experiment. 

\paragraph{One-ID} The pre-training model is fine-tuned by adding the user's unique ID to the beginning of the input prompt. In this paper, the user ID is a random four-digit alphanumeric string. Since user switching is done only by the user ID, only one model is trained with this method. This model is advantageous in terms of data volume because it can learn the combined dialogue history of all users, but it may also generate the utterances of other users, which may pose a risk in terms of privacy and security.

\paragraph{LoRA} User-specific fine-tuning models trained with LoRA are explained in \ref{subsec:LoRA}. One LoRA model was trained per user. We set the rank to $r=12$, and the number of trainable parameters was 1.8 million.

\paragraph{FULL} User-specific models with all pre-trained model parameters updated by fine-tuning. Each user has one full dialogue model.

\vspace{1em}

Before the fine-tuning, domain adaptation learning to the experimental data was performed on the pre-trained models. This is because the experimental data set described below is actual spoken dialogue and chat data, not reply pairs on which the pre-trained models were trained, and fine-tuning with a small amount of data is likely to learn corpus-derived styles rather than personal characteristics. For domain adaptation, we took all user halves from the training dataset and used the training data aggregated from them. Fine-tuning in the experiment used the other half of the data.

For fine-tuning, we used Adam \cite{2015-kingma} with a learning rate of $lr=1e^{-4}$. Early stopping was set to terminate learning when the validation loss began to rise.

\subsection{Experiment 1:Next utterance prediction} \label{sub:next_utt}

We first evaluated whether the dialogue models could predict the utterance in the corpus given user information and dialogue context. If a model can generate utterances close to actual utterances in the same context as the corpus is input, it can reproduce the user's dialogue. We evaluated utterance prediction by training user-specific dialogue models on multiple dialogue corpora.

\subsubsection{Experiment datasets}

\begin{table}[t]
    \centering
    \scriptsize
    \begin{tabular}{l|c|c|c}
    \hline
        dataset & SD & UPC & PCJP \\ \hline
        N.users & 147 & 140 & 100 \\
        user & Real & Real & Persona \\
        domain & speed dating & open  & open  \\
        style & speech conversation  & text chat & text chat \\
        N. all train &  \multicolumn{1}{r|}{180,703} & \multicolumn{1}{r|}{23,757} & \multicolumn{1}{r}{154,882} \\
        N.train / user & \multicolumn{1}{r|}{1,650} & \multicolumn{1}{r|}{168} & \multicolumn{1}{r}{1,548} \\
        N.test / user & \multicolumn{1}{r|}{282} & \multicolumn{1}{r|}{47} & \multicolumn{1}{r}{73} \\ \hline
    \end{tabular}
    \vspace{-1em}
    \caption{\footnotesize Experiment datasets. No. of train/user and No. of test/user are average numbers per user. }
    \label{table:eval_dataset}
\end{table}

We prepared three datasets with different properties for the experiments, as summarized in Table \ref{table:eval_dataset}.

\paragraph{Speed Dating (SD)} \citetlanguageresource{ishii:int2023} collected 147 face-to-face conversations between pairs of men and women who had never met. The conversations were about finding a romantic partner (speed dating). Each user has a detailed profile and self-introduction. Our experiment used a corpus of the collected conversations converted into text.

\paragraph{User Profile Chat (UPC)} The text chat data is from 140 users who met for the first time. Each user has detailed profile information and a self-introduction. During the chat, each user interacted with the other user by referring to the other user's profile information. The profile information does not contain any personally identifiable information (e.g., names or exact addresses) and was created with careful attention to privacy. Note that these interactive data are small amounts of data collected in a short period of about half a day.

\paragraph{PersonaChatJP (PCJP)} The Japanese version of the PERSONA-CHAT open corpus was created by \citetlanguageresource{10022973} \footnote{\url{https://github.com/nttcslab/japanese-dialog-transformers}}. It features 100 persona users who have five persona descriptions, and the dialogue is an open-domain chat about the persona descriptions. Since profile information such as gender and age are not specified in the persona information, the PPP and PSP enter the profile inferred from the persona information and utterances.

All of these datasets contain multiple interactions with other users. We extracted the results for 30\% of the dialogue partners and created a set with the extracted user's dialogues as development and test data and the rest as training data. We created three sets and used them to perform validation tests.

\subsubsection{Comparison models}

\paragraph{Fine-tuning based models} We trained user-specific models for all users in the evaluation datasets using the four pre-training models described in \ref{subsub:pretrain} and the three fine-tuning methods described in \ref{subsub:finetun}. For example, for a PCJP with 100 users, three fine-tunings with ID, LoRA, and FULL were trained for 100 users for one pre-trained model, resulting in 201 models. The total number of models was 804 for the four pre-trained models (Plain T5, Plain Dialogue, and PPP/PSP Dialogue), and three sets of validation were performed, resulting in 2,412 models to be tested. Then, 3,372 models were tested with UPC and 3,540 with SD.

\paragraph{Prompt-based models} Since the experimental data is in Japanese, the following two models, which are available in Japanese, were used for comparison.

\begin{description}
  \item[- GPT3.5] GPT3.5 \cite{NEURIPS2020_1457c0d6} is an LLM model running on OpenAI's ChatGPT service \cite{chatgpt2022}. 
  In this paper, we role-played a speaker to GPT3.5 and generated an utterance with instructions to output the next utterance in the dialogue history.
  The prompt describes the user profiles of the role-playing speaker and partner, a self-introductory sentence created by the speaker, and a dialogue history of up to 10 turns.
 We accessed  gpt-3.5-turbo\footnote{\url{https://platform.openai.com/docs/guides/gpt}} via OpenAI-API in September 2023.
  \item[- Jp-NeoX-3.6B]Jp-NeoX-3.6B is a Japanese GPT-NeoX model with 3.6 billion parameters \cite{rinnaneox2023}. This model is a text completion model. We used input prompts with the same content as GPT3.5 for the input prompt, modified into a completion format.
    \item[- Jp-Llama2-7B] Jp-Llama2-7B is a model based on Llama2 \cite{touvron2023llama} with additional pre-training to extend Japanese language proficiency \cite{elyzallama2023}.
    The input prompt uses the same method as Jp-NeoX-3.6B.
\end{description}

\begin{table*}[h]
\scriptsize
  \begin{minipage}[t]{.45\textwidth}
    \begin{center}
\begin{tabular}{r|rrrr}
\hline
 \multicolumn{5}{c}{Speed Dating (SD)} \\
 \hline
 & \multicolumn{1}{c}{Similarity} & \multicolumn{1}{c}{Acc@sim0.9} &\multicolumn{1}{c}{Rouge-L} & \multicolumn{1}{c}{PPL} \\
\hline
\multicolumn{1}{l}{Plain T5}  & \multicolumn{1}{l}{}  &\multicolumn{1}{l}{}   & \multicolumn{1}{l}{}  &  \\ 
\hline
One-ID & \textbf{0.549} & \textbf{0.0906} & \textbf{20.7} & 19.9\\
LoRA & 0.515 & 0.0556 & 17.8 & \textbf{12.7}\\
FULL & 0.518 & 0.0574 & 18.1 & 14.0 \\
\hline
\multicolumn{1}{l}{Plain Dialogue}  & \multicolumn{1}{l}{}  &\multicolumn{1}{l}{}   & \multicolumn{1}{l}{}  &  \\ 
\hline
One-ID &  \textbf{0.558} & \textbf{0.103} & \textbf{21.3} & 22.2\\
LoRA & 0.518 & 0.0612 & 18.1 & 24.3\\
FULL & 0.521 & 0.0647 & 18.4 & \textbf{17.2}\\
\hline
\multicolumn{1}{l}{PSP Dialogue}  & \multicolumn{1}{l}{}  &\multicolumn{1}{l}{}   & \multicolumn{1}{l}{}  &  \\ 
\hline
One-ID & \textbf{0.523} & \textbf{0.0620} & \textbf{18.6} & 16.1\\
LoRA & 0.505 & 0.0545 & 17.5 & 16.1\\
FULL & 0.510 & 0.0554 & 17.8 & \textbf{15.7}\\
\hline
\multicolumn{1}{l}{PPP Dialogue}  & \multicolumn{1}{l}{}  &\multicolumn{1}{l}{}   & \multicolumn{1}{l}{}  &  \\
\hline
One-ID & \textbf{0.566} & \textbf{0.110} & \textbf{22.1} & 21.5\\
LoRA & 0.518 & 0.0620 & 18.1 & 13.07\\
FULL & 0.520 & 0.0660 & 18.4 & \textbf{10.7}\\
\hline\hline
GPT3.5 & 0.267 & 0.0130 & 10.9 & NA  \\
Jp-NeoX-3.6B & 0.306 & 0.0218 & 7.37 & NA  \\
Jp-Llama2-7B & 0.278 & 0.0146 & 6.56 & NA  \\
\hline
\end{tabular}
    \vspace{-1em}
    \caption{\footnotesize Experiment results for next utterance prediction in speed dating dataset. Bold items indicate the highest value for each metric.}
    \label{table:result_sd}
    \end{center}
  \end{minipage}
  \hfill
  \begin{minipage}[t]{.45\textwidth}
    \begin{center}
\begin{tabular}{r|rrrr}
\hline
 \multicolumn{5}{c}{User Profile Chat (UPC)} \\
 \hline
 & \multicolumn{1}{c}{Similarity} & \multicolumn{1}{c}{Acc@sim0.9} &\multicolumn{1}{c}{Rouge-L} & \multicolumn{1}{c}{PPL} \\
\hline
\multicolumn{1}{l}{Plain T5}  & \multicolumn{1}{l}{}  &\multicolumn{1}{l}{}   & \multicolumn{1}{l}{}  &  \\ 
\hline
One-ID & \textbf{0.497} & \textbf{0.0467} & \textbf{23.7} & 23.7\\
LoRA & 0.443 & 0.0235 & 21.4 & \textbf{23.0}\\
FULL & 0.448 & 0.0282 & 22.2 & 23.3\\
\hline
\multicolumn{1}{l}{Plain Dialogue}  & \multicolumn{1}{l}{}  &\multicolumn{1}{l}{}   & \multicolumn{1}{l}{}  &  \\ 
\hline
One-ID & \textbf{0.530} & \textbf{0.0711} & 27.8 & 15.9\\
LoRA & 0.520 & 0.0691 & 27.8 & 16.6\\
FULL & 0.524 & 0.0727 & \textbf{28.2} & \textbf{14.8}\\
\hline
\multicolumn{1}{l}{PSP Dialogue}  & \multicolumn{1}{l}{}  &\multicolumn{1}{l}{}   & \multicolumn{1}{l}{}  &  \\ 
\hline
One-ID & 0.533 & 0.0721 & 27.9 & 17.2\\
LoRA & 0.532 & 0.0781 & 28.0 & 12.6\\
FULL & \textbf{0.533} & \textbf{0.079} & \textbf{28.2} & \textbf{10.1}\\
\hline
\multicolumn{1}{l}{PPP Dialogue}  & \multicolumn{1}{l}{}  &\multicolumn{1}{l}{}   & \multicolumn{1}{l}{}  &  \\
\hline
One-ID & 0.533 & 0.072 & 27.4 & 17.7\\
LoRA & 0.530 & 0.0758 & 27.9 & 14.5\\
FULL & \textbf{0.534} & \textbf{0.0780} & \textbf{28.0} & \textbf{14.4}\\
\hline\hline
GPT3.5 & 0.342 & 0.0447 & 18.4 & NA  \\
Jp-NeoX-3.6B & 0.317 & 0.0306 & 15.8 & NA  \\
Jp-Llama2-7B & 0.313 & 0.0373 & 15.1 & NA  \\
\hline
\end{tabular}
    \vspace{-1em}
    \caption{\footnotesize Experiment results for next utterance prediction in user profile chat dataset. Bold items indicate the highest value for each metric.}
    \label{table:result_ppc}
    \end{center}
  \end{minipage}
\end{table*}

\begin{table}[t]
\scriptsize
\begin{tabular}{r|rrrr}
\hline
 \multicolumn{5}{c}{PersonaChatJP (PCJP) } \\
 \hline
 & \multicolumn{1}{c}{Similarity} & \multicolumn{1}{c}{Acc@sim0.9} &\multicolumn{1}{c}{Rouge-L} & \multicolumn{1}{c}{PPL} \\
\hline
\multicolumn{1}{l}{Plain T5}  & \multicolumn{1}{l}{}  &\multicolumn{1}{l}{}   & \multicolumn{1}{l}{}  &  \\ 
\hline
One-ID & \textbf{0.567} & \textbf{0.0791} & \textbf{24.4} & \textbf{23.6}\\
LoRA & 0.556 & 0.0679 & 23.9 & 16.5\\
FULL & 0.556 & 0.0590 & 23.2 & 29.3\\
\hline
\multicolumn{1}{l}{Plain Dialogue}  & \multicolumn{1}{l}{}  &\multicolumn{1}{l}{}   & \multicolumn{1}{l}{}  &  \\ 
\hline
One-ID & \textbf{0.569} & \textbf{0.0775} & \textbf{25.4} & 22.2\\
LoRA & 0.560 & 0.0698 & 24.8 & \textbf{17.4}\\
FULL & 0.563 & 0.0681 & 24.5 & 19.2\\
\hline
\multicolumn{1}{l}{PSP Dialogue}  & \multicolumn{1}{l}{}  &\multicolumn{1}{l}{}   & \multicolumn{1}{l}{}  &  \\ 
\hline
One-ID & \textbf{0.570} & 0.0785 & 24.3 & 18.8\\
LoRA & 0.562 & 0.0769 & \textbf{25.2} & \textbf{14.7}\\
FULL & 0.566 & \textbf{0.0797} & 24.9 & 16.1\\
\hline
\multicolumn{1}{l}{PPP Dialogue}  & \multicolumn{1}{l}{}  &\multicolumn{1}{l}{}   & \multicolumn{1}{l}{}  &  \\
\hline
One-ID & 0.571 & \textbf{0.0805} & 25.3 & 17.8\\
LoRA & \textbf{0.572} & 0.0784 & \textbf{25.6} & \textbf{16.5}\\
FULL & 0.570 & 0.0767 & 24.7 & 17.9\\
\hline\hline
GPT3.5 & 0.350 & 0.0549 & 16.9 & NA  \\
Jp-NeoX-3.6B & 0.309 & 0.0162 & 14.3 & NA  \\
Jp-Llama2-7B & 0.331 & 0.0283 & 14.8 & NA  \\
\hline
\end{tabular}
\vspace{-1em}
    \caption{\footnotesize Experiment results for next utterance prediction in PersonaChatJP dataset. Bold items indicate the highest value for each metric.}
    \label{table:result_pcjp}
\end{table}

\begin{table}[t]
 \centering
 \scriptsize
 \begin{tabular}{l|c|c|c}
 \hline
 Model & One-ID & LoRA & FULL \\ \hline
 No. of models & 1 & No. of users & No. of users \\
 Model size & 850 MB & 7 MB & 850 MB \\
 Response time & 0.24 sec & 0.55 sec & 2.7 sec \\
 \hline
 \end{tabular}
 \vspace{-1em}
 \caption{\footnotesize Specifications of fine-tuning models. ``Response time'' means average time between loading user-specific model and generating utterance.}
 \label{table:model_spec}
\end{table}

\begin{table*}[t]
\begin{center}    
\scriptsize
\begin{tabular}{r|rrr|rrr|rrr}
\hline
 & \multicolumn{3}{c|}{Speed Dating (SD)} & \multicolumn{3}{c|}{User Profile Chat (UPC)} & \multicolumn{3}{c}{PersonaChatJP (PCJP) } \\
 \hline
 & \multicolumn{1}{c}{Dist-1} &\multicolumn{1}{c}{Dist-2} & \multicolumn{1}{c|}{dist-S} & \multicolumn{1}{c}{Dist-1} &\multicolumn{1}{c}{Dist-2} & \multicolumn{1}{c|}{Dist-S} & \multicolumn{1}{c}{Dist-1} &\multicolumn{1}{c}{Dist-2} & \multicolumn{1}{c}{Dist-S} \\
\hline
\multicolumn{1}{l}{Plain Dialogue}  & \multicolumn{1}{l}{}  &\multicolumn{1}{l}{}   & \multicolumn{1}{l}{}  &  \\ 
\hline
One-ID & 0.0253 & 0.0612 & 0.0411 &  0.174 & 0.320 & 0.305 & 0.240 & 0.529 & 0.840 \\
LoRA & 0.0204 & 0.0421 & 0.127 & 0.0330 & 0.0669 & 0.265 & 0.126 & 0.280 & 0.736 \\
FULL & 0.0264 & 0.0581 & 0.204  & 0.0348 & 0.0725 & 0.312 & 0.1317& 0.303 & 0.806  \\
\hline
\multicolumn{1}{l}{PSP Dialogue}  & \multicolumn{1}{l}{}  &\multicolumn{1}{l}{}   & \multicolumn{1}{l}{}  &  \\ 
\hline
One-ID &0.0310 & 0.0699 & 0.0547 &  0.0324 & 0.0624 & 0.181 &  0.163 & 0.364 & 0.826 \\
LoRA & 0.0253 & 0.0648 & 0.391 & 0.0457 & 0.103 & 0.472 &  0.139 & 0.304 & 0.763 \\
FULL &  0.0276 & 0.0721 & \textbf{0.451}  & 0.0443 & 0.100 & \textbf{0.475} & 0.141 & 0.322 & 0.794\\
\hline
\multicolumn{1}{l}{PPP Dialogue}  & \multicolumn{1}{l}{}  &\multicolumn{1}{l}{}   & \multicolumn{1}{l}{}  &  \\
\hline
One-ID & 0.0314 & 0.0713 & 0.0410 & 0.0206 & 0.0354 & 0.0877 & 0.154 & 0.347 & 0.753\\
LoRA & 0.0316 & 0.0759 & 0.345  & 0.0496 & 0.109 & 0.452 & 0.160 & 0.356 & 0.837 \\
FULL & \textbf{0.0372} & \textbf{0.0924} & 0.418  & \textbf{0.0517} & \textbf{0.112} & 0.456 & \textbf{0.161} & \textbf{0.374} & \textbf{0.885} \\
\hline
\end{tabular}
    \vspace{-1em}
    \caption{\footnotesize Experiment results for diverse utterance generation. Bold items indicate the highest value for each metric.
    Dist-1 and  Dist-2 represent uni-grams and bi-grams; Dist-S is an extension of Distinct-N to utterance sentence units.}
    \label{table:result_div}
\end{center}
\end{table*}

\subsubsection{Metrics}

We used embedding-based metrics \cite{liu-etal-2016-evaluate} to evaluate the similarity of the test data to the reference utterances. The embeddings were created using SentenceBERT \cite{reimers-gurevych-2019-sentence}\footnote{\url{https://huggingface.co/sonoisa/sentence-luke-japanese-base-lite}}, and cosine distance was used for similarity. We also introduced Acc@sim0,9, which indicates the percentage of generated utterances nearly identical to the reference utterance (i.e., where the similarity of embedding a generated utterance and a reference utterance is greater than or equal to 0.9). We also used Rouge-L, which evaluates the longest sequence length, and perplexity, which evaluates the performance of the language model.

\subsubsection{Result}

The results are listed in Tables \ref{table:result_sd}, \ref{table:result_ppc}, and \ref{table:result_pcjp}. The results of each model are discussed in detail in the following paragraphs.

\paragraph{LoRA/FULL vs One-ID} When PlainT5 and Plain Dialogue were used as pre-training models, the One-ID model scored higher than the other fine-tuning methods. On the other hand, in the case of the PSP or PPP pre-training models that add user profiles, the models with LoRA/FULL user-specific models scored slightly higher, especially on the UPC or PCJP datasets with fewer training data.

The results show that it is difficult to learn user-specific models from PlainT5 or Plain Dialogue models. This is likely because the user's utterance generation, learned through fine-tuning, is very different from the pre-trained model. As a likely result, more training data was needed to learn sufficiently. In other words, by including prompts based on the user's profile during pre-training, the user-specific model can be additionally trained using the dialogue tendencies of users close to the training target by inputting a profile similar to that of the individual model user when fine-tuning the individual model, so that even with a small amount of training data, the training could reflect the individuality of the user.

\paragraph{LoRA/FULL vs Prompt} The reproducibility of utterance generation using GPT3.5 and other LLMs with user profiles included in the prompts scored lower than models trained individually with user utterances. This suggests that LLMs make reasonable responses to the context of prompt information that includes a user's profile and dialogue history but that these responses are not necessarily the user's specific responses.

\begin{figure}[t]
\centering
\includegraphics[width =.45\textwidth]{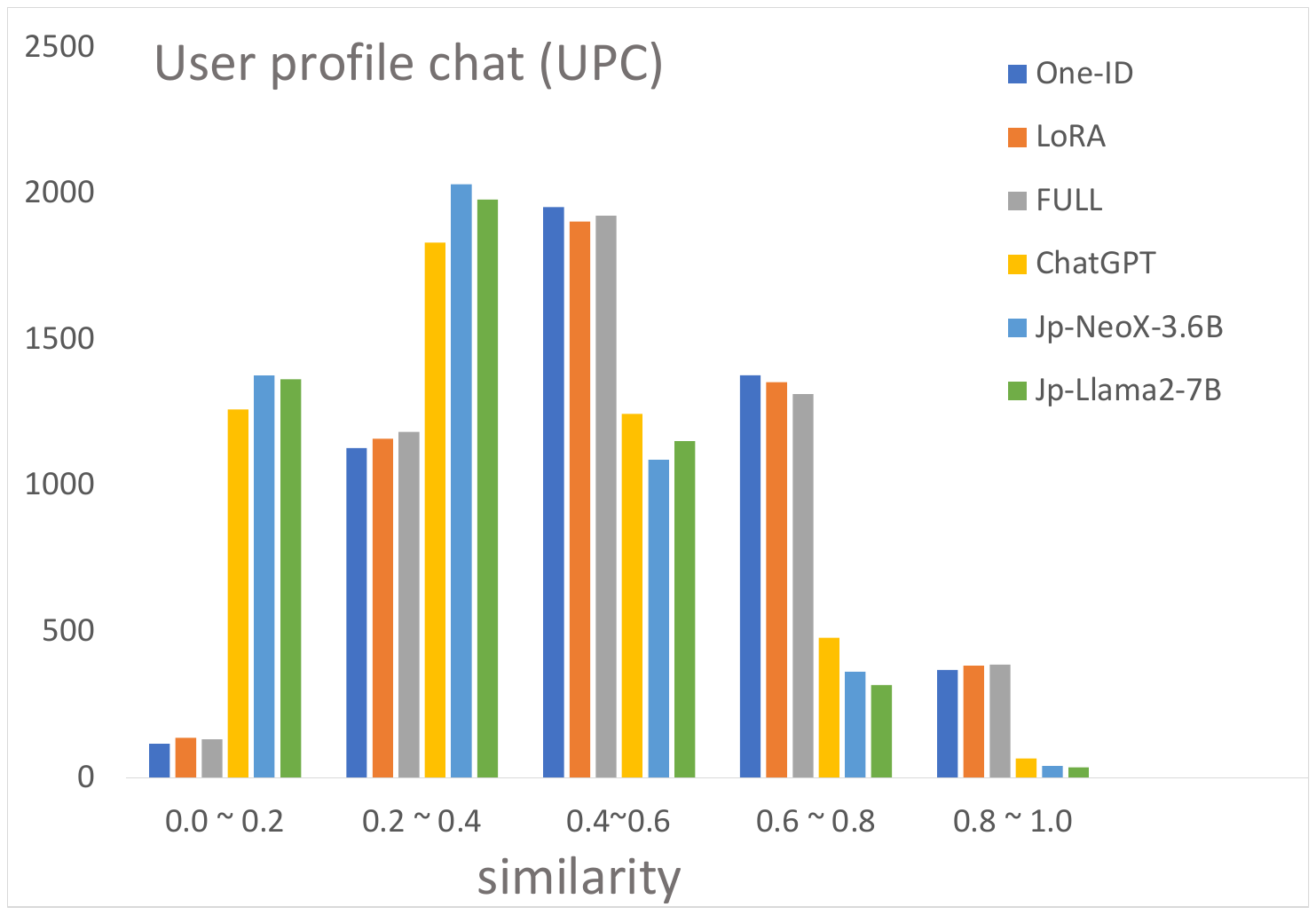}
\vspace{-0.5em}
\caption{\footnotesize Frequency distribution of similarity between generated and reference utterances for user profile chat dataset.}
\label{fig:exp1_models}
\end{figure}

We show in Fig. \ref{fig:exp1_models} a graph of the frequency of similarity between generated and reference utterances, arranged in 0.2 increments. Prompt-based utterance generation produced a large number of utterances with a similarity of under 0.2 to the reference utterances. This means that the generated utterances were almost entirely different from the reference utterances. The fine-tuned user-specific models could generate more reproducible utterances, even though the model sizes were much smaller than the compared LLMs. These results suggest that real user interaction, unlike fictional characters or role-playing, is challenging to reproduce with prompts alone, and fine-tuning is essential even for small data sets.

\paragraph{LoRA vs FULL} Both LoRA and FULL are fine-tuned user-specific models for pre-training. The results show that FULL, which updates all parameters, generated slightly more reproducible utterances. However, no significant differences between the two were observed.

LoRA and FULL should be chosen according to the application. Table \ref{table:model_spec} shows the specifications of the fine-tuning models used in this experiment. The FULL model requires one complete model per user. The model used in the experiment was a relatively small model with 222 million parameters, which is not a significant problem when training a model dedicated to a few persons. However, when the number of users increases, it consumes more resources. On the other hand, LoRA only has a small-sized adapter model, so even if the number of users with dedicated models increases, the resource consumption would be minimal compared to FULL. The small model size also makes it possible to dynamically switch adapters while generating utterances, allowing for large-scale scaling.

\subsection{Experiment 2: Diverse utterance generation} 
\label{sub:div_utt}
We evaluated the reproducibility of the dialogue corpus in \ref{sub:next_utt}. We define the task in this paper as using dialog context, user information, and user ID to determine utterances. Even if the input dialogue context is the same, the utterance must change depending on the user's information. In the experiments in this section, we evaluated the diversity of utterances produced when different user information was input for the same dialogue context.

\subsubsection{Experiment datasets}
We used 70 questions from \citetlanguageresource{10.1007/978-3-319-09767-1_53}'s personality question set for dialogue context. 

\subsubsection{Comparison models}
We compared three models learned in the \ref{sub:next_utt} experiment, precisely, the One-ID and the LoRA/FULL models, which showed high performances in the \ref{sub:next_utt} experiments. The inference algorithm used a greedy search to avoid the influence of the sampling probability.

\subsubsection{Metrics}
Since this experiment evaluated the diversity of generated utterances, we used the Distinct-N metric proposed by Li et al. \cite{li-etal-2016-diversity}. Distinct-N calculates the proportion of different N-grams. For the same input question, we computed the Distinct-N of the set of utterances generated by all user models, with a higher Distinct-N indicating more N-grams for the same question. In other words, it shows us how various utterances can be generated depending on the users.

\subsubsection{Results}

The results for the diversity of user utterances are listed in Table \ref{table:result_div}. The One-ID model had lower diversities of produced utterances than LoRA/FULL. Although the One-ID model switches users by the ID of the input prompt in one model, the expression of switching users may be weak. This means that it could generate learned utterances from other users, which could be problematic from a security and privacy perspective.

The model that generated the most variety of utterances was the FULL model fine-tuned from a pre-trained model using PPP, followed by the LoRA models. FULL and LoRA's difference was similar to the utterance reproducibility experiment described in \ref{sub:next_utt}. LoRA and FULL have the advantage of not learning other users' utterances because they train user-specific models on user-specific data.

\section{Conclusion}

This paper described a method for learning dialogue models to achieve user-specific personalized dialogue. We proposed a method combining a small pre-trained dialogue model that includes a simple user profile prompt in the input with a user-specific fine-tuning model using parameter-efficient fine-tuning. Experimental evaluations using comparative models that combine multiple datasets, pre-training models, and fine-tuning demonstrated the effectiveness of the proposed method. Furthermore, we conducted personality reproduction experiments with prompt-based generated utterances using the proposed model and large language models (LLMs). We found that the proposed method of fine-tuning with a small amount of training data reproduces personality better than utterances generated from LLMs with prompts containing personal features.

In the future, we will work on speech generation based on personal memories, which is essential for speech demonstrating the individuality of real users. We also plan to proceed with verification of the effectiveness of this method when used with more detailed prompts with LLMs.

\nocite{*}
\section{Bibliographical References}\label{sec:reference}

\bibliographystyle{lrec-coling2024-natbib}
\bibliography{lrec-coling2024-example}

\section{Language Resource References}
\label{lr:ref}
\bibliographystylelanguageresource{lrec-coling2024-natbib}
\bibliographylanguageresource{languageresource}

\end{document}